\documentclass[letterpaper]{article} 
\usepackage{aaai21}  
\usepackage{times}  
\usepackage{helvet} 
\usepackage{courier}  
\usepackage[hyphens]{url}  
\usepackage{graphicx} 
\usepackage{natbib}  
\usepackage{caption} 
\usepackage[switch]{lineno}  %

\usepackage{enumitem}
\usepackage{multirow}
\usepackage{soul,color}

\usepackage{subfig}

\title{Hostility Detection Dataset in Hindi}
\author {
Mohit Bhardwaj$^{\dagger}$, Md Shad Akhtar$^{\dagger}$, Asif Ekbal$^{\ddagger}$, Amitava Das$^{\star}$, Tanmoy Chakraborty$^{\dagger}$
\\
}

\affiliations {
$^{\dagger}$\textit{IIIT Delhi}, India. 
$^{\ddagger}$\textit{IIT Patna}, India. 
$^{\star}$\textit{Wipro Research}, India. \\

{\tt \{mohit19014,tanmoy,shad.akhtar\}@iiitd.ac.in,  asif@iitp.ac.in, amitava.das2@wipro.com 
}
}

\begin{document}
\maketitle

\begin{abstract}
    In this paper, we present a novel hostility detection dataset in Hindi language. We collect and manually annotate $\sim8200$ online posts. The annotated dataset covers four hostility dimensions: fake news, hate speech, offensive, and defamation posts, along with a non-hostile label. The hostile posts are also considered for multi-label tags due to a significant overlap among the hostile classes. We release this dataset as part of the CONSTRAINT-2021 shared task on hostile post detection. 
\end{abstract}

\section{Introduction}
The COVID-19 pandemic has changed our lives forever, both online and offline. As the physical world went into lockdown, the online world came closer than ever. Since people are confined to their homes, they are spending way more time on social media, chat rooms, communication apps, and gaming servers, which can have serious implications on the mental health on an individual as well. 
According to a recent survey\footnote{\url{https://l1ght.com/Toxicity_during_coronavirus_Report-L1ght.pdf}}, there has been 900\% increase in hate speech towards China and it's people on Twitter, 200\% increase in traffic on sites that promote hate speech against Asians, 70\% increase in hate speech among teen and kids online, and toxicity levels in gaming community has increased by 40\% as well. Similar trends have been observed in non-English languages as well, and since the percentage of non-English tweets in India\footnote{\url{https://bit.ly/3mnXoKM}} have risen up to 50\%, early detection of hostile texts in low resource languages like Hindi is of paramount importance as well.

A significant number of online social media users post harmful contents without realising that they are crossing the line defined by the freedom-of-speech. As evident as it sounds, fake news, hate speeches, offensive remarks, etc., are extremely harmful for any civilised society, and ask for adequate and appropriate guidelines to prevent/curb such activities. The foremost task in neutralising such activities is the hostile post detection, and many works have been carried out to address the issue in English \cite{Waseem2016HatefulSO,Waseem2017UnderstandingAA,Nobata2016AbusiveLD}. 

Despite Hindi being the third most spoken language in the world, and a significant presence of Hindi content on social media platforms, to our surprise, we were not able to find any significant dataset on fake news or hate speech detection in Hindi. A survey of the literature suggest a few works related to hostile post detection in Hindi, such as \cite{kar2020rumours,jha:et:al:2020:offfensive:hindi,safi-samghabadi-etal-2020-aggression}; however, there are two basic issues with these works - either the number of samples in the dataset are not adequate or they cater to a specific dimension of the hostility only. In this paper, we present our manually annotated dataset for hostile posts detection in Hindi. We collect more than $\sim8200$ online social media posts and annotate them as hostile and non-hostile posts. Furthermore, we identify four hostility dimensions for each hostile post as \textit{fake}, \textit{defamation}, \textit{hate}, and \textit{offensive}. Though some of these hostile dimensions sound similar at the abstract-level (e.g., \textit{hate} and \textit{offensive}), their definitions are different, and we define them below following \cite{Mathur2018DidYO} and \cite{Davidson2017AutomatedHS}.

\begin{itemize}
    \item \textbf{\textit{Fake News}}: A claim or information that is verified to be not true. We have included tweets belonging to click bait and satire/parody categories as fake news as well.
    
    \item \textbf{\textit{Hate Speech}}: A post targeting a specific group of people based on their ethnicity, religious beliefs, geographical belonging, race, etc., with malicious intentions of spreading hate or encouraging violence. 
    
    \item \textbf{\textit{Offensive}}: A post containing profanity, impolite, rude, or vulgar language to insult a targeted individual or group.
    
    \item \textbf{\textit{Defamation}}: A mis-information regarding an individual or group, which is destroying their reputation publicly.
    
    \item \textbf{\textit{Non-Hostile}}: A post with no hostility.
\end{itemize}

The dataset development is part of the CONSTRAINT-2021 shared task \cite{constraints:shared:tasks}. The CONSTRAINT-2021 workshop emphasizes the hostility detection on three major points, i.e., low-resource regional languages, detection in emergency situations, and early detection.  Currently, the train and validation set is available with the shared task\footnote{\url{https://constraint-shared-task-2021.github.io/}}, and we will release the test set at the end of the workshop. 

\section{Related work}
Waseem \& Hovy \cite{Waseem2016HatefulSO} considered annotations for hate speech, but they did not consider other dimensions of hostile text like offensive or bullying. In another work, Waseem et al. \cite{Waseem2017UnderstandingAA} discuss the user agreement and consensus in annotating bullying, harassment, offensive, and hate speeches. They showed that it is easy to identify the victim of the bully quite convincingly by the annotators, whereas there is very low consensus in annotations of harassment, offensive and hate speech. This may be partially because hostility can be generalized, directed, implicit or explicit. Wijesiriwardene et al. \cite{wijesiriwardene2020alone} provide a dataset of toxicity (harassment, offensive language, hate speech) on Twitter in English.  

An example of implicit hostility in Hindi is to call someone `\textit{meetha}', which literally means \textit{sweet} in Hindi; however, the intended meaning in a hostile post could be of `\textit{fa\$\$ot}' - a derogatory term towards the LGBT community.

Among other notable works on hostility detection, Davidson et al. \cite{Davidson2017AutomatedHS} studied the hate speech detection for English. They argued that some words may reflect hate in one region; however, the same word can be used as a frequent slang term. For example, in English, the term `\textit{dog}' does not reveal any hate or offense, but it (\textit{ku\#\#a}) is commonly referred to as a derogatory term in Hindi. 

Considering the severity of the problem, some effort has been made for non-English languages as well, such as Arabic \cite{Haddad2020ArabicOL}, Bengali \cite{Hossain2020BanFakeNewsAD}, Hindi \cite{jha:et:al:2020:offfensive:hindi}, etc. Samghabadi et al. \cite{safi-samghabadi-etal-2020-aggression} addressed the problem of aggression and misogyny detection in English, Hindi, and Bengali, whereas, Jha et al. \cite{jha:et:al:2020:offfensive:hindi} worked on the keyword-based (swear words) offensive text detection in Hindi. There are also a few attempts at Hindi-English code-mixed hate speech \cite{Bohra2018ADO} and offensive post \cite{Mathur2018DetectingOT} detection. Recently, Kar et al. \cite{kar2020rumours} developed a multi-lingual COVID-19 rumour detection dataset in English, Hindi, and Bangla; however, the dataset is significantly small and is limited to COVID-related text only.   

\begin{figure}[t!]
\centering
\subfloat[Hostile Word Cloud. \label{fig:word:cloud:hostile}]{
    \includegraphics[width=0.22\textwidth]{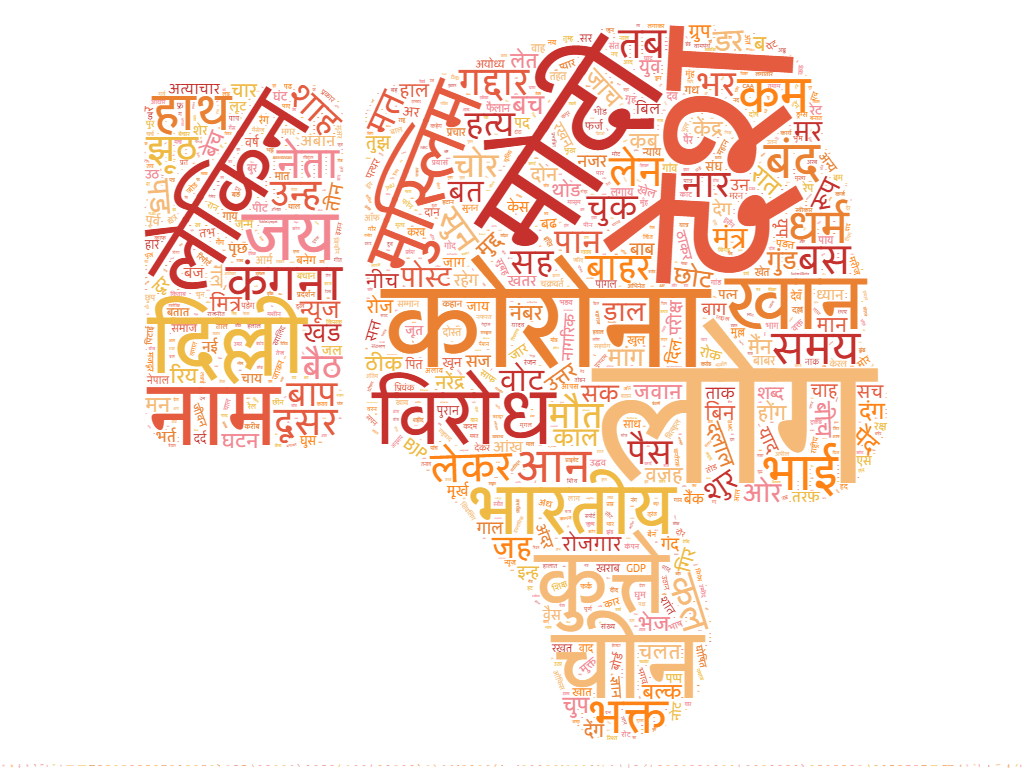}
    }
\subfloat[Non-Hostile Word Cloud.\label{fig:word:cloud:non:hostile}]{
    \includegraphics[width=0.22\textwidth]{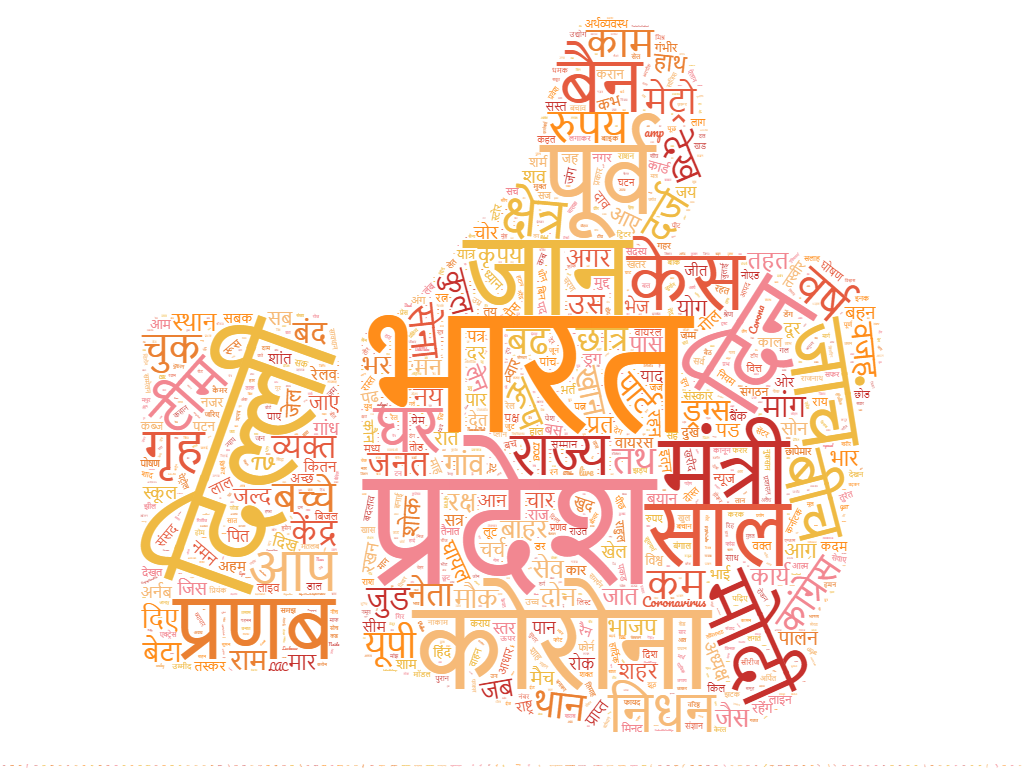}
    }
\caption{Word clouds.}
\label{fig:word:cloud}
\end{figure}

\begin{table*}[ht!]
    \centering
 \resizebox{\textwidth}{!}{
    \begin{tabular}{l}
    \includegraphics[width=\textwidth]{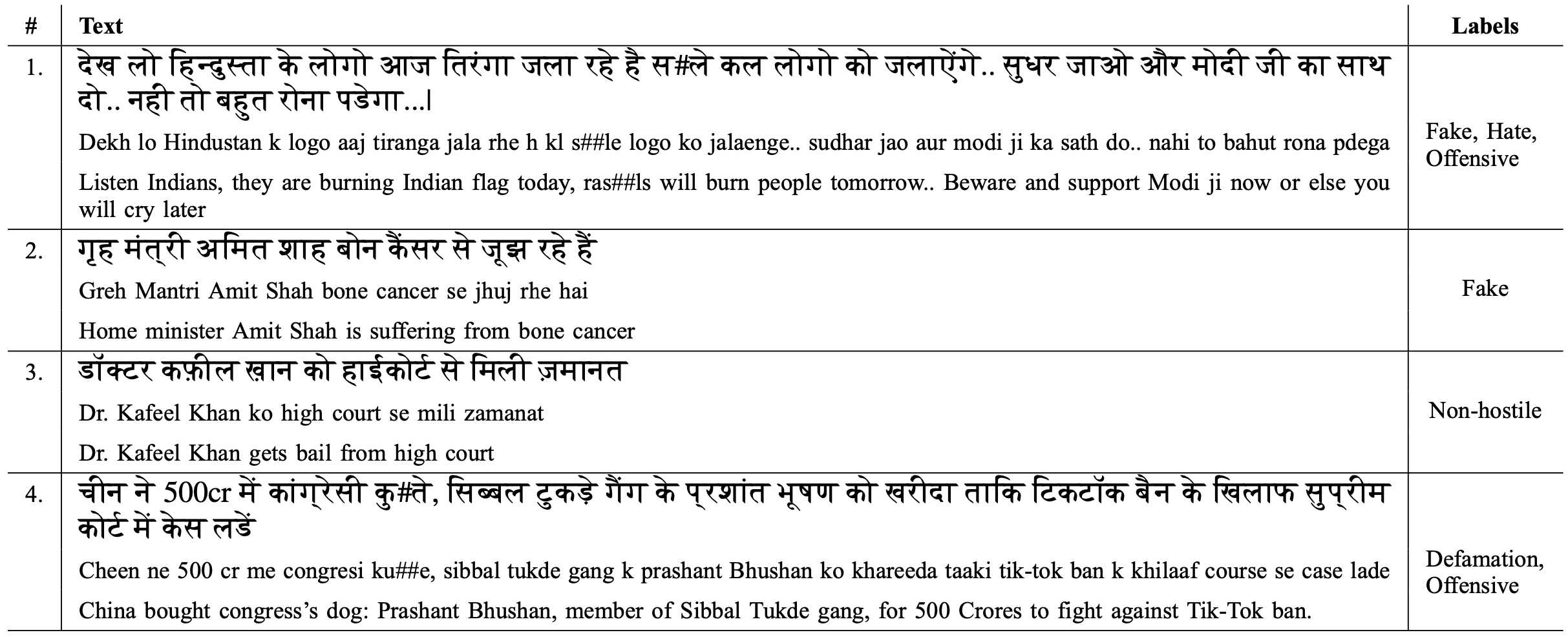} \\
    \end{tabular}}
    \caption{A few annotated examples from the dataset.}
     \label{tab:examples}
\end{table*}

\section{Data Development}
\begin{figure}[t!]
\centering
\includegraphics[width=0.3\textwidth]{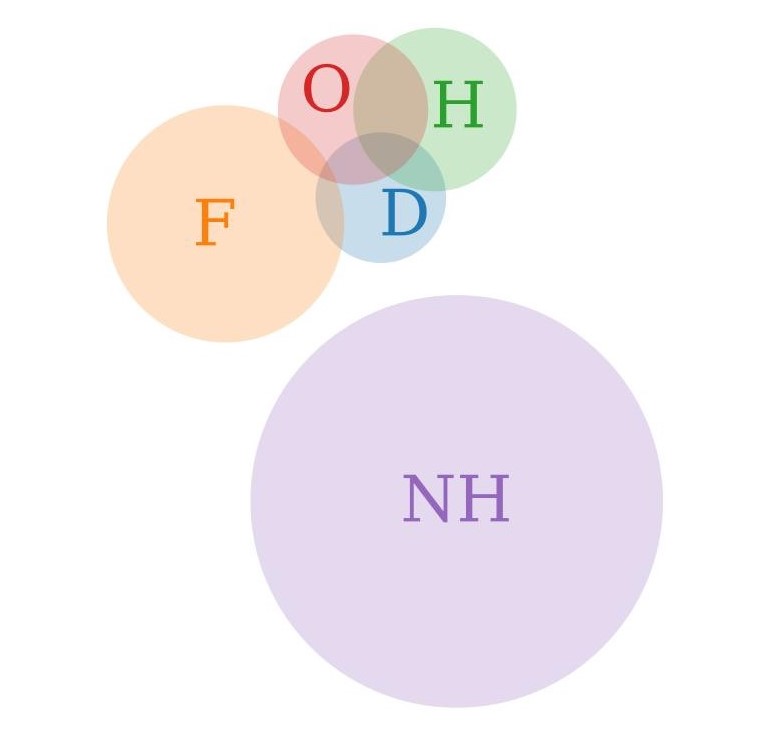}
\caption{Venn Diagram of Multi-Label Hindi hostility dataset. Notations: [$F-$ \textit{Fake}], [$O-$ \textit{Offensive}], [$H-$ \textit{Hate}], [$D-$ \textit{Defamation}], [$NH-$ \textit{Non-hostile}].}
\label{fig:venn:diagram}
\end{figure}

During the development of the dataset, we observe that some of the posts have overlap among the hostility dimensions; therefore, we adopt the idea of multi-label tagging for each post. Figure \ref{fig:venn:diagram} shows the class-wise overlaps among hostile dimensions in form of a Venn diagram\footnote{\url{https://www.meta-chart.com/venn}}. Although it reveals the relationship among for the majority cases, it is inadequate to show the intersection between fake and hate class in 2D. 

Some of the examples from the dataset are presented in Table \ref{tab:examples}. For example, the first sentence is a factually verified fake news, and lies in the offensive category due to the use of vulgar language, but it has an implicit hatred against religious minority, which might even lead to violence amongst two communities in the worst scenarios. Similarly in the fourth example, derogatory and vulgar language is used towards a famous advocate alongside spreading misinformation to defame him.

\subsection{Data Collection}
We collect $\sim 8200$ hostile and non-hostile texts from various social media platforms like Twitter, Facebook, WhatsApp, etc. We follow different strategies to collect data for each category.

\begin{itemize}[leftmargin=*]
    \item For \textit{fake news} collection, we refer to some of India's top most fact checking websites like BoomLive \footnote{{https://hindi.boomlive.in/fake-news}}, Dainik Bhaskar \footnote{{https://www.bhaskar.com/no-fake-news/}}, etc, and read numerous articles in Hindi. This process helps us identify the topics of the fake news. Subsequently, we compile a topic-wise keyword list for each fake news. Next, we curate online social media platforms such as Twitter, Instagram, etc., for the collection of posts. 

    \item For \textit{hate speech} collection, at first, we target the tweets encouraging violence against minorities based on their race, religious beliefs, etc. Following this process, we analyse the timelines of users with significant hate-related posts. Additionally, we also analyse users who liked or commented in support of the hate speech and scan their timelines for additional hate-related posts as well.

    \item For \textit{offensive posts}, we employ the list of top swear words used in Hindi language as determined by Jha et al. \citep{jha:et:al:2020:offfensive:hindi}. For each swear word, we query Twitter API\footnote{{https://developer.twitter.com/en/docs/twitter-api}} to extract (offensive) tweets. In the next step, we manually verify each collected tweet as offensive. One critical observation that we make during the collection process is that offensive posts against women are more toxic and hate-oriented than the male counterpart. 
    
    \item For the posts related to the \textit{defamation} category, we read viral news articles where people or a group are publicly shamed due to misinformation, and perform topic-wise search to collect defamation tweets.
    
    \item To collect \textit{non-hostile} data, we extract posts from some of the trusted sources (e.g., BBCHindi). We manually iterate over the collected samples to ensure that they are non-hostile in every way. Furthermore, we also annotate around 600 non-hostile texts from many non verified users with small followers count to maintain diversity in our dataset.
\end{itemize}

\subsection{Dataset Details}
A brief statistics of the dataset is presented in Table \ref{tab:data_stat}. Out of $8192$ online posts, $4358$ samples belong to non-hostile category, while the rest $3834$ posts convey one or more hostile dimensions. In the annotated dataset, there are 1638, 1132, 1071, and 810 posts for \textit{fake}, \textit{hate}, \textit{offensive}, and \textit{defame} classes, respectively. Note that each post can belong to multiple hostile dimensions as depicted in Table \ref{tab:examples}. We split the dataset into 80:10:20 for train, validation, and test, by ensuring the uniform label distribution among the three sets, respectively.

\begin{figure*}[t!]
\centering
\subfloat[Average number of characters and words per post. \label{fig:average:words}]{
    \includegraphics[scale=0.28]{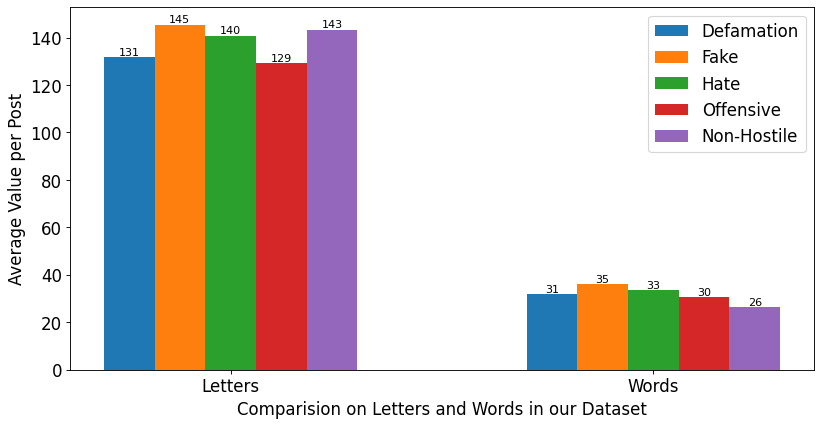}
    }
\hspace{2em}
\subfloat[Average Punctuations (| ,  : ? \- \_ " ; !), Hashtags, and User Mentions per post.\label{fig:average:pucnt}]{
    \includegraphics[scale=0.28]{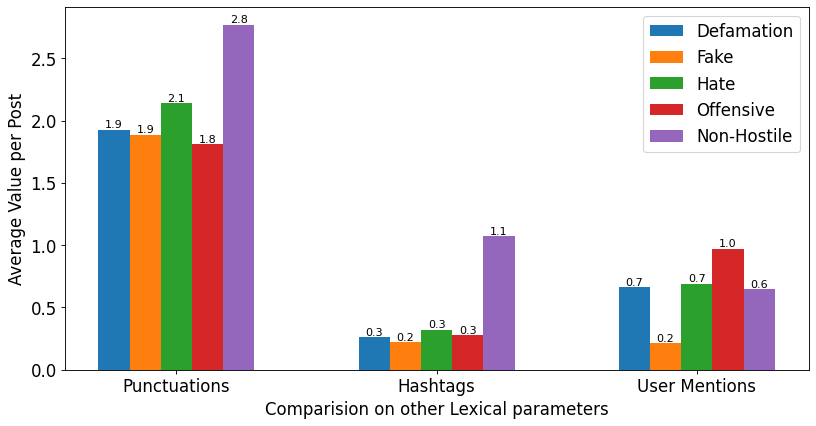}
}
\caption{Class-wise distribution.}
\label{fig:classwise:distri}
\end{figure*}

On analyzing our dataset, we find multiple interesting patterns. Figure \ref{fig:average:words} shows the average number of letters and words per post across each hostile and non-hostile dimension. It is interesting to note that unlike other languages, in Hindi even though the hostile posts have a higher average number of letters per post, the average number of words in hostile posts is lower than the non-hostile posts. Similarly from Figure \ref{fig:average:pucnt}, we can observe that non-hostile posts in Hindi have $32\%$ more punctuation marks on average when compared to hostile posts. This might depict the lack of concern for correct grammar while someone is being hostile towards another person. Another interesting feature to note is this that offensive hostile category has a single user mention in the post on average, thus suggesting a directed offensive posts in our dataset. 

We also show the word clouds\footnote{\url{https://www.wordclouds.com/}} in hostile and non-hostile posts in Figures \ref{fig:word:cloud:hostile} and \ref{fig:word:cloud:non:hostile}, respectively.
There are some common popular words which belong to both hostile and non-hostile categories. This is because words like Corona, Modi, Nation, and many more were over social media throughout our entire annotation process in all sorts of conversations. Still, the amount of negation and offensiveness against the ruling party, against Muslims, or countries like China, Pakistan is clearly visible in Figure \ref{fig:word:cloud:hostile}. 

\begin{table}[t!]
    \centering
    \resizebox{0.48\textwidth}{!}{
    \begin{tabular}{l|c|c|c|c|c|c}
    & \multicolumn{5}{c|}{\bf Hotile posts} & \multirow{2}{*}{\bf Non-hostile} \\ \cline{2-6}
    & \bf Fake & \bf Hate & \bf Offense & \bf Defame & \bf Total$^*$ & \\ \hline
    
    \hline
         Train      & 1144 & 792  & 742  & 564 & 2678 & 3050 \\
         Validation & 160  & 103  & 110  & 77  & 376  & 435  \\
         Test       & 334  & 237  & 219  & 169 & 780  & 873  \\ 
         Overall    & 1638 & 1132 & 1071 & 810 & 3834 & 4358 \\ \hline
    \end{tabular}}
        \caption{Dataset statistics and Label distribution. Fake, hate, defame, and offense reflect the number of respective posts including multi-label cases. $^*$ denotes total hostile posts.}
        \label{tab:data_stat}
\end{table}

\begin{table}[t!]
    \centering
    \resizebox{0.48\textwidth}{!}{
    \begin{tabular}{c|c|c|c|c|c|c}
          &  & \multicolumn{1}{c|}{\multirow{2}{3em}{Coarse grained}} & \multicolumn{4}{c}{Fine grained} \\ \cline{4-7}
         Model & Embedding & & Hate & Fake & Offensive & Defamation \\\hline \hline
         
          LR & \multirow{4}{*}{m-BERT} & 83.98 & 44.27 & \textbf{68.15} & 38.76 & 36.27 \\
         SVM &                        & \textbf{84.11} & \textbf{47.49} & 66.44 & \textbf{41.98} & \textbf{43.57}\\
         RF &                         & 79.79 & 6.83 & 53.43 & 7.01 & 2.56\\
         MLP &                        & 83.45 & 34.82 & 66.03 & 40.69 & 29.41\\ 
         
         \hline
    \end{tabular}}
    \caption{Coarse grained and Fine grained results (F1 score) of various models. }
    \label{tab:results}
\end{table}

\section{Evaluation}
We benchmark our dataset employing four traditional machine learning algorithms, i.e., Support Vector Machine (SVM), Decision Tree (DT), Random Forest (RF), and Logistic Regression (LR). We evaluate our model on the validation set\footnote{Please note that the test set has not been released yet. It will be released soon} and report weighted-F1 scores for each case. 

Prior to training the models, we perform a few pre-processing steps as follows: 
\begin{itemize}[leftmargin=*]
    \item \textbf{Stopwords:} Removal of stopwords following the list available at Data Mendeley \footnote{\url{https://data.mendeley.com/datasets/bsr3frvvjc/1}}
    \item \textbf{Non-Alphanumeric Characters: } We remove all other non-alphanumeric character except full stop punctuation marks ('|') in Hindi.
    \item \textbf{Emojis:} For convenience, we skip emojis, emoticons, symbols, pictographs, transport, maps, dingbats, flags, etc.
    \item \textbf{URLs:} We also remove all URLs from the post if any.
\end{itemize}

\subsection{Models and Implementation Details}
We employ the multilingual BERT\footnote{\url{https://huggingface.co/bert-base-multilingual-uncased}} (m-BERT) pre-trained \cite{devlin2019bert} model for computing the input embedding. We extract the last layer of the pre-trained model as the corresponding word embedding for each word in a sentence. Further, we represent the sentence as the average of constituents' embedding.

Employing the sentence embeddings as our features, we trained models for both coarse-grained and fine-grained tasks. We employ Scikit-learn library for the implementation with following hyper-parameters. For SVM, we use a linear kernel with $c=0.01$ and $\gamma=1$. For Random Forests, we set the number of estimators as 400, while for Logistic Regression, we use `$l1$' penalty with `\textit{liblinear}' solver. In case of MLP, we define two hidden layers having 30 and 10 neurons each, followed by a softmax layer. We use `Relu' as the activation function and a learning rate of 0.001. For each case, the rest of the parameters are default.

We use one vs all strategy for training all our fine-grained models. We use our hostile samples only to train our fine-grained models for each hostility dimension. This is done to reduce the data imbalance, as we have around 800 samples on average for training in any hostile dimension. In SVM and Logistic Regression, we use the class\_weight parameter as 'balanced', which allows the model to find the right weights on it's for imbalance classes.

\begin{figure}[ht!]
 \centering
 \subfloat[SVM \label{fig:confusion:svm}]{
  \includegraphics[width=0.23\textwidth]{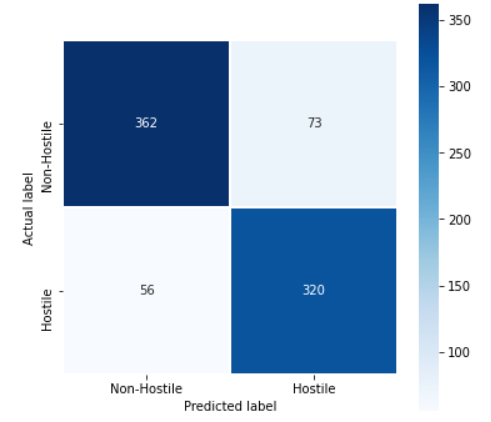}}
  \subfloat[Random-Forest \label{fig:confusion:rf}]{  \includegraphics[width=0.23\textwidth]{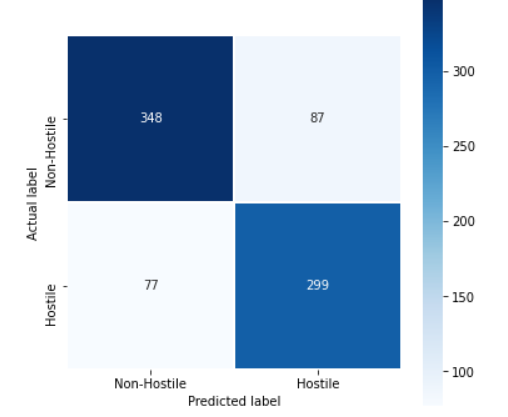}}

\subfloat[MLP\label{fig:confusion:mlp}]{  
\includegraphics[width=0.23\textwidth]{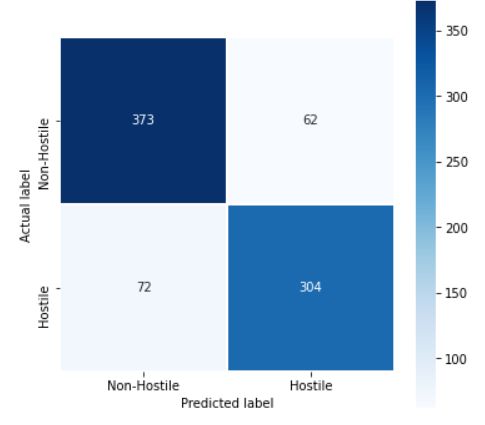}}
  \subfloat[Logistic Regression \label{fig:confusion:lr}]{  
\includegraphics[width=0.23\textwidth]{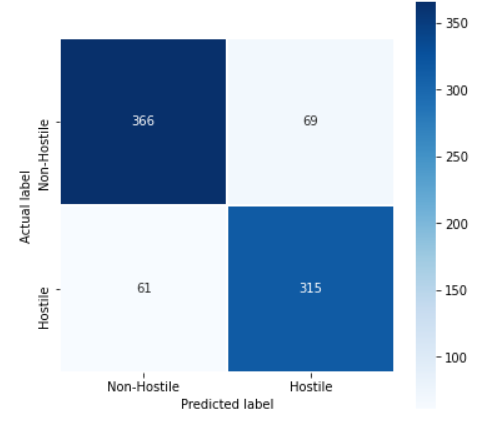}}
\caption{Confusion matrices of ML algorithms on Hostility dataset.}
\label{fig:confusion}
\end{figure}

\section{Results}
We use the validation set for evaluating the performance of our models and report the obtained results in Table \ref{tab:results}. In coarse-grained evaluation, SVM reported the best weighted-F1 score of 84.11\%, whereas, we obtain 83.98\%, 83.45\%, and 79.79\% w-F1 scores for LR, MLP, and RF, respectively. For each case, we present the confusion matrix in Figure \ref{fig:confusion}. 
In fine-grained evaluation, SVM reports the best F1-score for three hostile dimensions, i.e., Hate (47.49\%), Offensive (41.98\%), and Defamation (43.57\% ), whereas, Logistic Regression outperforms others in \textit{Fake} dimension with F1-score of 68.15\%. 

\section{Conclusion}
In this paper, we present the development process of a novel, multi-dimensional hostility detection dataset in Hindi. We manually annotated $\sim8200$ posts as hostile or non-hostile across various social media platforms. Furthermore, we assigned fine-grained hostile labels to each hostile post, i.e., \textit{fake}, \textit{hate}, \textit{offensive}, and \textit{defamation}. We also provide four baseline systems to benchmark our dataset. 

\bibliography{references}

\begin{thebibliography}{15}
\providecommand{\natexlab}[1]{#1}
\providecommand{\url}[1]{\texttt{#1}}
\providecommand{\urlprefix}{URL }
\expandafter\ifx\csname urlstyle\endcsname\relax
  \providecommand{\doi}[1]{doi:\discretionary{}{}{}#1}\else
  \providecommand{\doi}{doi:\discretionary{}{}{}\begingroup
  \urlstyle{rm}\Url}\fi

\bibitem[{con(2021)}]{constraints:shared:tasks}
 2021.
\newblock {CONSTRAINTS-2021: Shared tasks on Hostile Posts Detection}.
\newblock \urlprefix\url{http://lcs2.iiitd.edu.in/CONSTRAINT-2021/}.

\bibitem[{Bohra et~al.(2018)Bohra, Vijay, Singh, Akhtar, and
  Shrivastava}]{Bohra2018ADO}
Bohra, A.; Vijay, D.; Singh, V.; Akhtar, S.; and Shrivastava, M. 2018.
\newblock A Dataset of Hindi-English Code-Mixed Social Media Text for Hate
  Speech Detection.
\newblock In \emph{PEOPLES@NAACL-HTL}.

\bibitem[{Davidson et~al.(2017)Davidson, Warmsley, Macy, and
  Weber}]{Davidson2017AutomatedHS}
Davidson, T.; Warmsley, D.; Macy, M.; and Weber, I. 2017.
\newblock Automated Hate Speech Detection and the Problem of Offensive
  Language.
\newblock In \emph{ICWSM}.

\bibitem[{Devlin et~al.(2019)Devlin, Chang, Lee, and
  Toutanova}]{devlin2019bert}
Devlin, J.; Chang, M.-W.; Lee, K.; and Toutanova, K. 2019.
\newblock {BERT}: Pre-training of Deep Bidirectional Transformers for Language
  Understanding.
\newblock In \emph{Proceedings of the 2019 Conference of the North {A}merican
  Chapter of the Association for Computational Linguistics: Human Language
  Technologies, Volume 1 (Long and Short Papers)}, 4171--4186. Minneapolis,
  Minnesota: Association for Computational Linguistics.
\newblock \doi{10.18653/v1/N19-1423}.
\newblock \urlprefix\url{https://www.aclweb.org/anthology/N19-1423}.

\bibitem[{Haddad et~al.(2020)Haddad, Orabe, Al-Abood, and
  Ghneim}]{Haddad2020ArabicOL}
Haddad, B.; Orabe, Z.; Al-Abood, A.; and Ghneim, N. 2020.
\newblock Arabic Offensive Language Detection with Attention-based Deep Neural
  Networks.
\newblock In \emph{OSACT}.

\bibitem[{Hossain et~al.(2020)Hossain, Rahman, Islam, and
  Kar}]{Hossain2020BanFakeNewsAD}
Hossain, M.~Z.; Rahman, M.~A.; Islam, M.~S.; and Kar, S. 2020.
\newblock BanFakeNews: A Dataset for Detecting Fake News in Bangla.
\newblock In \emph{LREC}.

\bibitem[{Jha et~al.(2020)Jha, Poroli, N, Vijayan, and
  P}]{jha:et:al:2020:offfensive:hindi}
Jha, V.; Poroli, H.; N, V.; Vijayan, V.; and P, P. 2020.
\newblock DHOT-Repository and Classification of Offensive Tweets in the Hindi
  Language.
\newblock \emph{Procedia Computer Science} 171: 2324--2333.
\newblock \doi{10.1016/j.procs.2020.04.252}.

\bibitem[{Kar et~al.(2020)Kar, Bhardwaj, Samanta, and Azad}]{kar2020rumours}
Kar, D.; Bhardwaj, M.; Samanta, S.; and Azad, A.~P. 2020.
\newblock No Rumours Please! A Multi-Indic-Lingual Approach for COVID
  Fake-Tweet Detection.
\newblock \emph{ArXiv} 2010.06906.

\bibitem[{Mathur et~al.(2018{\natexlab{a}})Mathur, Sawhney, Ayyar, and
  Shah}]{Mathur2018DidYO}
Mathur, P.; Sawhney, R.; Ayyar, M.; and Shah, R. 2018{\natexlab{a}}.
\newblock Did you offend me? Classification of Offensive Tweets in Hinglish
  Language.
\newblock In \emph{ALW}.

\bibitem[{Mathur et~al.(2018{\natexlab{b}})Mathur, Shah, Sawhney, and
  Mahata}]{Mathur2018DetectingOT}
Mathur, P.; Shah, R.; Sawhney, R.; and Mahata, D. 2018{\natexlab{b}}.
\newblock Detecting Offensive Tweets in Hindi-English Code-Switched Language.
\newblock In \emph{SocialNLP@ACL}.

\bibitem[{Nobata et~al.(2016)Nobata, Tetreault, Thomas, Mehdad, and
  Chang}]{Nobata2016AbusiveLD}
Nobata, C.; Tetreault, J.; Thomas, A.; Mehdad, Y.; and Chang, Y. 2016.
\newblock Abusive Language Detection in Online User Content.
\newblock In \emph{WWW}.

\bibitem[{Safi~Samghabadi et~al.(2020)Safi~Samghabadi, Patwa, PYKL, Mukherjee,
  Das, and Solorio}]{safi-samghabadi-etal-2020-aggression}
Safi~Samghabadi, N.; Patwa, P.; PYKL, S.; Mukherjee, P.; Das, A.; and Solorio,
  T. 2020.
\newblock Aggression and Misogyny Detection using {BERT}: A Multi-Task
  Approach.
\newblock In \emph{Proceedings of the Second Workshop on Trolling, Aggression
  and Cyberbullying}, 126--131. Marseille, France: European Language Resources
  Association (ELRA).
\newblock ISBN 979-10-95546-56-6.
\newblock \urlprefix\url{https://www.aclweb.org/anthology/2020.trac-1.20}.

\bibitem[{Waseem et~al.(2017)Waseem, Davidson, Warmsley, and
  Weber}]{Waseem2017UnderstandingAA}
Waseem, Z.; Davidson, T.; Warmsley, D.; and Weber, I. 2017.
\newblock Understanding Abuse: A Typology of Abusive Language Detection
  Subtasks.
\newblock \emph{ArXiv} abs/1705.09899.

\bibitem[{Waseem and Hovy(2016)}]{Waseem2016HatefulSO}
Waseem, Z.; and Hovy, D. 2016.
\newblock Hateful Symbols or Hateful People? Predictive Features for Hate
  Speech Detection on Twitter.
\newblock In \emph{SRW@HLT-NAACL}.

\bibitem[{Wijesiriwardene et~al.(2020)Wijesiriwardene, Inan, Kursuncu, Gaur,
  Shalin, Thirunarayan, Sheth, and Arpinar}]{wijesiriwardene2020alone}
Wijesiriwardene, T.; Inan, H.; Kursuncu, U.; Gaur, M.; Shalin, V.~L.;
  Thirunarayan, K.; Sheth, A.; and Arpinar, I.~B. 2020.
\newblock ALONE: A Dataset for Toxic Behavior Among Adolescents on Twitter.
\newblock In Aref, S.; Bontcheva, K.; Braghieri, M.; Dignum, F.; Giannotti, F.;
  Grisolia, F.; and Pedreschi, D., eds., \emph{Social Informatics}, 427--439.
  Cham: Springer International Publishing.
\newblock ISBN 978-3-030-60975-7.

\end{thebibliography}
\end{document}